\documentclass[conference]{IEEEtran}
\IEEEoverridecommandlockouts

\usepackage{cite}
\usepackage{amsmath,amssymb,amsfonts}
\usepackage{algorithm}
\usepackage{algorithmic}
\usepackage{graphicx}
\usepackage{textcomp}
\usepackage{xcolor}

\usepackage[utf8]{inputenc}
\usepackage{comment}
\usepackage{booktabs}
\usepackage{subcaption}
\usepackage{multirow}
\usepackage{placeins}

\def\BibTeX{{\rm B\kern-.05em{\sc i\kern-.025em b}\kern-.08em
    T\kern-.1667em\lower.7ex\hbox{E}\kern-.125emX}}

\newcommand{\calD}{\mathcal{D}}
\newcommand{\calL}{\mathcal{L}}
\newcommand{\calE}{\mathcal{E}}
\newcommand{\calA}{\mathcal{A}}
\newcommand{\R}{\mathbb{R}}

\begin{document}

\title{FLEX-MoE: Federated Mixture-of-Experts with Load-balanced Expert Assignment for Edge Computing}

\author{\IEEEauthorblockN{\begin{tabular}{c}
Boyang Zhang$^{1}$, Xiaobing Chen$^{1}$, Songyang Zhang$^{2}$, Shuai Zhang$^{3}$, Xiangwei Zhou$^{1}$,\\
Jian Zhang$^{1}$, Mingxuan Sun$^{1}$
\end{tabular}}
\IEEEauthorblockA{\begin{tabular}{c}
$^{1}$Louisiana State University: \{bzhan29,xchen87,xwzhou,jz,msun11\}@lsu.edu\\
$^{2}$University of Louisiana at Lafayette: songyang.zhang@louisiana.edu\\
$^{3}$New Jersey Institute of Technology: sz457@njit.edu\\
\end{tabular}}
}

\maketitle

\begin{abstract}

Mixture-of-Experts (MoE) models enable scalable neural networks through conditional computation, offering enhanced effectiveness and efficiency for next-generation wireless communications. However, deploying MoE with federated learning (FL) over wireless and IoT edge networks faces two critical challenges: 1) resource-constrained clients cannot store large AI models with full expert sets, and 2) non-IID data distributions cause severe expert load imbalance that degrades model performance. To this end, we propose \textbf{FLEX-MoE}, a federated MoE framework that jointly optimizes expert assignment and load balancing under limited client capacity. Specifically, our approach introduces client-expert fitness scores that quantify expert suitability for local datasets through training feedback, and employs an optimization-based algorithm to maximize client-expert specialization while enforcing balanced expert utilization system-wide. Unlike greedy methods that focus solely on personalization while ignoring load imbalance, FLEX-MoE addresses expert utilization skew, which is particularly severe in heterogeneous edge FL. Our experimental results demonstrate superior accuracy and consistently balanced expert utilization across diverse resource-constrained scenarios for edge computing.


\end{abstract}

\section{Introduction}

Federated Learning (FL) and Mixture-of-Experts (MoE) architectures have emerged as powerful paradigms to enable efficient and specialized training of large-scale AI models across wireless sensor networks and Internet-of-Things (IoT). %
FL provides a decentralized collaborative learning framework, 
which ensures privacy protection and reduces communication overhead \cite{mcmahan2017communication,10557587}. On the other hand, MoE architectures \cite{shazeer2017outrageously,fedus2022switch,chen2023sparse,chen2026efficient}
partition large neural networks into specialized experts, which are conditionally activated based on input data, thereby enabling significant scaling of model parameters while maintaining computational efficiency. Recent studies on MoE for wireless communications further show the potential of placing experts across base stations, mobile devices, and edge nodes, where expert selection must account for communication latency, channel conditions, and device capability \cite{xue2024wdmoe,song2025mixture}. 
In this work, we will investigate the federated MoE for next-generation wireless communications and edge computing.

Unlike conventional centralized MoE, federated MoE over wireless communication networks faces significant resource constraints arising from edge clients with limited computational resources, where devices such as smartphones, unmanned aerial vehicles (UAVs), and IoT nodes exhibit highly variable memory, processing power, bandwidth, and access latency \cite{liu2024split,lee2024recurrent,10589575}. These computational constraints fundamentally prohibit clients from synchronizing and maintaining all %
the
experts locally, creating a critical scalability bottleneck for the deployment of federated MoE \cite{mei2024fedmoe}. 
However, existing \textit{full expert storage methods} in distributed MoE implicitly assume abundant client resources, requiring clients to maintain all experts locally \cite{reisser2021federated,guo2021pfl,tran2025revisiting}, which fundamentally limits their practical deployment in resource-constrained environments.

To address limited computational resources, a straightforward solution is to assign experts randomly based on client capacity. However, such an approach overlooks data heterogeneity and fails to exploit the specialized data patterns that are unique to each client. In particular, data heterogeneity, resulting from non-IID data distributions among clients, poses unique challenges for expert specialization in federated training of MoE \cite{lu2024federated,dun2023fedjets,yi2024pfedmoe}. Unlike traditional FL with a single dense model, distributed MoE architectures require careful assignment of specialized experts to match the characteristics of local data, as different clients benefit from vastly different combinations of experts for optimal personalization \cite{luo2024mixture}.

Although some recent work has investigated expert selection in federated MoE settings, most efforts have focused on client personalization without addressing imbalanced expert utilization, which is a key challenge for scalability and generalization.
In traditional MoE settings, imbalanced expert utilization is already a known issue that can degrade performance \cite{shazeer2017outrageously,fedus2022switch,wang2024auxiliary}. This problem can be significantly amplified in federated MoE, as data heterogeneity across clients can lead to vastly different expert preferences, potentially exacerbating the inherent expert utilization skew from MoE routing mechanisms. 
For example, \textit{greedy top-K assignment methods} assign experts based on scoring metrics \cite{dun2023fedjets,mei2024fedmoe} while ignoring expert load balance among clients. Other federated MoE variants target narrow settings such as prompt experts, domain-aware aggregation, or fixed two-expert models \cite{luo2024mixture,zhan2024fedmoe,yi2024pfedmoe}, while full-storage methods still require each client to maintain all experts \cite{tran2025revisiting}.
Unlike methods solely emphasizing personalized model performance for individual clients, we aim to federatedly train a unified MoE model where all experts are sufficiently engaged during training, thereby enhancing the generalization and learning effectiveness.

To address the aforementioned challenges, we propose \textbf{F}ederated \textbf{MoE} with \textbf{L}oad-balanced \textbf{EX}pert assignment (\textbf{FLEX-MoE}), a novel framework that dynamically assigns experts in a way that is capable of balancing expert utilization and enabling effective specialization under strict resource constraints. Our approach operates through a server-coordinated assignment strategy: the server dynamically collects client resource capacities and computes a newly proposed client-expert fitness score based on training feedback to quantify the suitability between clients and experts. 
Leveraging these fitness scores, we design an expert assignment algorithm based on linear programming to jointly maximize client-expert specialization quality while ensuring balanced expert utilization across clients. Unlike existing methods that require clients to maintain all experts or use greedy top-K selection without load balance, clients in FLEX-MoE synchronize their own assigned expert subsets based on individual capacity constraints, employ locally trained gating networks, and participate in server-coordinated expert aggregation to maintain both global model coherence and balanced expert training. More specifically, our key contributions can be summarized as follows:

\begin{itemize}
\item We address the novel problem of \textbf{joint expert assignment and load balancing under resource constraints} in federated MoE for edge computing, proposing FLEX-MoE, a framework that optimally allocates experts to resource-constrained clients while ensuring balanced expert utilization across the system. To the best of our knowledge, this is the first work to explicitly address expert load imbalance in federated MoE training.

\item We introduce a novel \textbf{client-expert fitness score} that quantitatively measures the suitability of experts for each client's local dataset by incorporating feedback metrics, enabling tailored expert selection that significantly improves personalization and training efficiency.

\item We design a \textbf{load-balanced expert assignment algorithm} that jointly optimizes client-expert specialization and system-wide load balancing under resource constraints. Unlike existing greedy approaches that focus solely on personalization, our method explicitly enforces balanced expert utilization to prevent expert uneven training and maintain model quality.

\item 
Our FLEX-MoE consistently delivers satisfactory accuracy while achieving the best load balance
compared to existing approaches. Notably, our experimental results further show that \textbf{enforcing load balance is essential for performance} under severely skewed data, where our method significantly outperforms all other strategies in both accuracy and load balance.

\end{itemize}

\begin{figure}[!t]
    \centering
    \includegraphics[width=0.98\columnwidth]{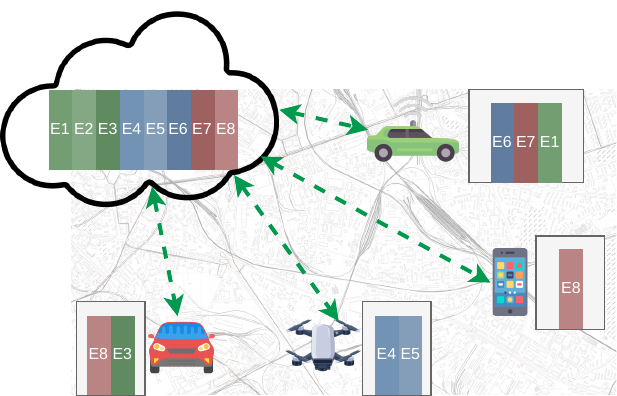}
    \caption{An edge intelligence scenario for federated MoE with resource constraints: heterogeneous wireless/IoT clients such as vehicles, UAVs, and mobile devices train different \textbf{expert subsets} under local capacity and communication constraints.}
    \label{fig:edge_scenario}
\end{figure}

\section{Problem Formulation}

In this work, we consider an FL system with a central server and $C$ clients indexed by $c \in \{1, ..., C\}$. Each client $c$ possesses a local dataset $\calD_c$ following non-IID distributions across clients, which cannot be shared with the server and other clients. An illustrative example is a UAV-assisted smart-city sensing system, where clients such as drones, roadside cameras, and traffic sensors collect signals under different wireless links and battery budgets. The goal is to collaboratively train a global MoE model with limited computational resources in a wireless/IoT system, where each client can only deploy a subset of experts as shown in Fig. \ref{fig:edge_scenario}. For simplicity, we focus on a single MoE layer within a larger architecture, (e.g., replacing a feed-forward block in a ResNet).
The MoE layer consists of $E$ experts, denoted by $\{\mathcal{E}_e\}_{e=1}^E$, each with parameters $\theta_e$. Each client $c$ maintains a personalized gating network $G_c$ with parameters $\phi_c$, which is designed to route inputs based on local data distributions.

The key challenges in federated MoE training include:
\begin{itemize}
    \item \textbf{Resource constraints.} Client $c$ can download and update at most $k_c \le E$ experts in any round due to computational limitations. We denote this round-specific assignment by a set $\mathcal{A}_c^{(t)}\subseteq\{1,\dots,E\}$ with $|\mathcal{A}_c^{(t)}| = k_c$.
    \item \textbf{Data heterogeneity.} Local datasets $\mathcal{D}_c$ are generally non-IID, requiring personalized expert assignment for effective specialization.
    \item \textbf{Expert load imbalance.} Popular experts may be over-selected while others remain underutilized, leading to degraded model performance.
\end{itemize}

The learning objective is to jointly optimize the global expert parameters $\{\theta_e\}_{e=1}^E$ and the local gating parameters $\{\phi_c\}_{c=1}^C$ to minimize the global loss function, defined as a weighted sum over all client data, i.e., 

\small
\begin{equation}
 \min_{\{\theta_e\}, \{\phi_c\}} \sum_{c=1}^C \frac{|\calD_c|}{|\calD|} \calL_c(\{\theta_e\}_{e \in \mathcal{A}_c^{(t)}}, \phi_c), 
\end{equation}
\normalsize
where $\calL_c$ is the average loss on client $c$'s local data $\calD_c$, $\mathcal{D} = \bigcup_{c=1}^C \mathcal{D}_c$, and $|\calD| = \sum_{c=1}^C |\calD_c|$.

The central problem addressed in this work is: \textit{how should the server assign a subset of $k_c$ experts to each client $c$ in each communication round $t$ to jointly (i) respect individual client capacity $k_c$, (ii) ensure experts assigned to client $c$ suit its local data $\calD_c$ for effective specialization, and (iii) keep all experts $\{\calE_e\}_{e=1}^E$ trained in a balanced manner across clients to prevent underutilization and preserve model quality?}

\section{ Method: FLEX-MoE}

We propose FLEX-MoE, a novel server-driven framework that addresses the joint optimization problem of expert assignment under resource constraints. Our approach dynamically assigns experts to clients based on client-expert fitness scores while enforcing system-wide load balancing. The framework continuously updates assignment decisions based on client feedback to enable resource-constrained, load-balanced, and specialization-aligned training.

\subsection{Overall Workflow}
\begin{figure}[h]
    \centering
    \includegraphics[width=0.95\columnwidth]{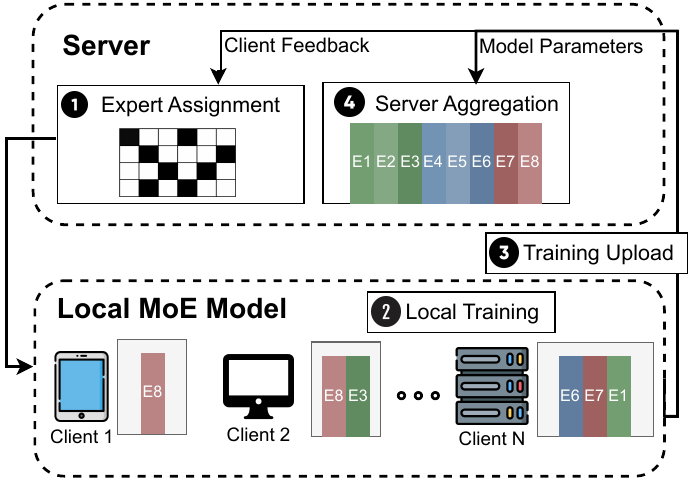}
    \caption{System-level workflow of FLEX-MoE: (1) Server assigns experts using optimization-based algorithm; (2) Clients perform local training with assigned experts; (3) Clients upload parameter updates and feedback; (4) Server aggregates updates.}
    \label{fig:diagram_fedmoe}
\end{figure}

We illustrate the FLEX-MoE training loop in Fig.~\ref{fig:diagram_fedmoe}. 
Each communication round $t$ ($t \ge 1$) proceeds through four phases, assuming the server has initialized $\{\theta_e^{0}\}_{e=1}^E$ and collected client capacities $\{k_c\}_{c=1}^{C}$. Only Steps 1 and 4 run on the server, while Steps 2 and 3 run on the client:

\begin{enumerate}    
    \item \textbf{Step 1: Server-side expert assignment.} At the beginning of round $t$, using client feedback (such as model accuracies received from all clients at the end of the previous round $(t-1)$ and the prior assignment $\mathcal{A}_c^{(t-1)}$ made in that round), the server computes a table of client-expert quality scores (detailed in Section \ref{sec:client_expert_matrix}) and employs an optimization-based assignment algorithm to determine the subset of expert indices $\calA_c^{(t)} \subseteq \{1, ..., E\}$ for each participating client $c$ for round $t$, ensuring $|\calA_c^{(t)}| = k_c$.
    
    \item \textbf{Step 2: Local training.}
    Upon receiving $\{\theta_e^{(t-1)}\}_{e\in\mathcal{A}_c^{(t)}}$ and the shared extractor $\psi^{(t-1)}$, client $c$ pairs them with its private gating network $G_c$ parameterized by~$\phi_c^{(t-1)}$ and performs $R$ local training rounds on $\mathcal{D}_c$. For image classification, each input $x_i$ is processed by $\psi^{(t-1)}$ to yield a feature map, $G_c$ routes it to experts in $\calA_c^{(t)}$, the gated expert outputs are aggregated and passed through the remaining layers to compute the loss $l(x_i,y_i;\Theta)$ (where $\Theta=(\psi,\phi_c,\{\theta_{c,e}\}_{e\in\calA_c^{(t)}})$), and a standard optimizer (e.g., Adam) updates all trainable parameters. At the end of local training, the client computes parameter updates $\Delta\psi_c^{(t)}$ and $\Delta\theta_{c,e}^{(t)}$ for every $e\in\mathcal{A}_c^{(t)}$.

    During local training, the client also records feedback for each assigned expert $e \in \calA_c^{(t)}$ in the current round $t$, such as the data counts, accuracy, and loss. Specifically, $u_{c,e}^{(t)}$ is the total count of data samples routed through expert $e$ by $G_c$ during the $R$ local training rounds. Let $a_{c,e}^{(t)}$ denote the local model accuracy (e.g., classification accuracy). Let $\ell_{c,e}^{(t)}$ denote the average loss calculated over the subset of images $\mathcal{D}_{c,e}^{(t)}$ routed to expert $e$ in the final local epoch. Specifically, it is calculated as
            \small
            \begin{equation}
            \ell_{c,e}^{(t)} = \frac{1}{|\mathcal{D}_{c,e}^{(t)}|} \sum_{(x_i, y_i) \in \mathcal{D}_{c,e}^{(t)}} l(x_i, y_i; \Theta).
            \label{eq:loss}
            \end{equation}
            \normalsize

    \item \textbf{Step 3: Client-side training upload.} After local training for round $t$, client $c$ uploads the parameter updates for the shared feature extractor $\Delta\psi_c^{(t)}$ and for the experts $\{\Delta\theta_{c,e}^{(t)}\}_{e \in \calA_c^{(t)}}$, along with client feedback such as accuracy and loss. The updated local gating parameters $\phi_c^{(t)}$ are retained by the client for the next round and are typically not sent to the server.

    \item \textbf{Step 4: Server aggregation.} Let $S_t$ be the set of participating clients. The shared extractor is averaged via FedAvg, $\Delta\psi^{(t)}=\sum_{c\in S_t}\frac{|\mathcal{D}_c|}{|\mathcal{D}^{(t)}|}\Delta\psi_c^{(t)}$, with $|\mathcal{D}^{(t)}|=\sum_{c\in S_t}|\mathcal{D}_c|$. Each expert $e$ is updated by usage-weighted averaging over clients assigned to it:
    {\small
    \begin{equation}
    \Delta\theta_e^{(t)} =
    \begin{cases}
    \frac{\sum_{c: e \in \calA_c^{(t)}} u_{c,e}^{(t)} \Delta\theta_{c,e}^{(t)}}{\sum_{c: e \in \calA_c^{(t)}} u_{c,e}^{(t)}} & \text{if } \sum_{c: e \in \calA_c^{(t)}} u_{c,e}^{(t)} > 0, \\
    0 & \text{otherwise}.
    \end{cases}
    \end{equation}}
    The server then sets $\psi^{(t)}=\psi^{(t-1)}+\Delta\psi^{(t)}$ and $\theta_e^{(t)}=\theta_e^{(t-1)}+\Delta\theta_e^{(t)}$, and retains feedback $\{u,a,\ell\}_{c,e}^{(t)}$ for the next round's assignment.

\end{enumerate}

\begin{algorithm}[h]
\small
\caption{FLEX-MoE Training Loop}
\label{alg:FLEX-MoE}
\begin{algorithmic}[1]
\REQUIRE Clients $C$, experts $E$, capacities $\{k_c\}$, rounds $T_{\text{max}}$
\STATE Server initializes $\{\theta_e^{(0)}\}_{e=1}^E$ and shared $\psi^{(0)}$
\STATE Each client $c$ initializes local gating network $G_c$ with $\phi_c^{(0)}$
\FOR{each round $t = 1$ to $T_{\text{max}}$}
    \STATE \textbf{Server-side Expert Assignment}
    \FOR{each client $c$}
        \STATE Update fitness $s_{c,e}^{(t-1)} = a_{c,e}^{(t-1)}$ or $\exp(-\alpha_L \ell_{c,e}^{(t-1)})$ via Eq.~\eqref{eq:indicator_acc} or~\eqref{eq:indicator_loss}
        \STATE $Q_t(c,e) = (1-\beta)\, Q_{t-1}(c,e) + \beta\, s_{c,e}^{(t-1)}$ via Eq.~\eqref{eq:update_q}
    \ENDFOR
    \STATE Compute target expert load: $\tau^{(t)} = (\sum_{c=1}^C |\mathcal{D}_c|\, k_c)/E$
    \STATE Construct dynamic bounds $L_{\text{new}}(e), \Gamma_{\text{new}}(e)$ for each expert $e$
    \STATE \textbf{Solve:} $\max \sum_{c,e} Q_{t-1}(c, e)\, X_{c,e}$ subject to $\sum_e X_{c,e}=k_c$ and $L_{\text{new}}(e) \le \sum_c |\mathcal{D}_c| X_{c,e} \le \Gamma_{\text{new}}(e)$
    \STATE Obtain assignment $\mathcal{A}_c^{(t)} = \{e \mid X_{c,e}=1\}$
    \STATE \textbf{Client-side Local Training}
    \FOR{each client $c \in S_t$}
        \STATE Download $\{\theta_e^{(t-1)}\}_{e\in\mathcal{A}_c^{(t)}}$ and $\psi^{(t-1)}$; train $R$ rounds on $\mathcal{D}_c$
        \STATE Upload $\Delta\psi_c^{(t)}$, $\{\Delta\theta_{c,e}^{(t)}\}$, and feedback $\{u,a,\ell\}_{c,e}^{(t)}$
    \ENDFOR
    \STATE \textbf{Server-side Aggregation}
    \STATE $\psi^{(t)} \gets \psi^{(t-1)} + \sum_{c\in S_t} \frac{|\mathcal{D}_c|}{|\mathcal{D}^{(t)}|}\Delta\psi_c^{(t)}$; aggregate each $\theta_e^{(t)}$ by usage-weighted averaging over assigned clients
\ENDFOR
\end{algorithmic}
\end{algorithm}

\subsection{Client-expert Quality Score Metrics} \label{sec:client_expert_matrix}

We now introduce the aforementioned client-expert quality score for expert assignment.
\subsubsection{Client-expert fitness Score (Q)} \label{sec:quality}

The server maintains a matrix $Q \in \R^{C \times E}$, where $Q(c, e)$ estimates the suitability or ``fitness'' of expert $e$ for client $c$'s local data. A higher score $Q(c,e)$ indicates a better fit (e.g., higher accuracy achieved or lower loss incurred by client $c$ with expert $e$). After each communication round $t$, for each client $c$ and each expert $e \in \mathcal{A}_c^{(t)}$ that it trained, the server updates $Q(c, e)$. This update is based on feedback from client $c$, such as the average local accuracy $a_{c,e}^{(t)}$ and loss $\ell_{c,e}^{(t)}$ observed for data samples processed by expert $e$ during local training at $t$.

The fitness indicator $s_{c,e}^{(t)}$ can be designed flexibly. In this work, we adopt two straightforward strategies:
\begin{itemize}
    \item \textbf{Accuracy Driven (Acc-driven):} The indicator is directly proportional to the reported accuracy:
    \small
    \begin{equation}
        s_{c,e}^{(t)} = a_{c,e}^{(t)}.
        \label{eq:indicator_acc}
    \end{equation}
    \normalsize
    \item \textbf{Loss Driven (Loss-driven):} The indicator is an exponential function of the negative loss, transforming lower loss values into higher scores:
    \small
    \begin{equation}
        s_{c,e}^{(t)} = \exp(-\alpha_L \ell_{c,e}^{(t)}),
        \label{eq:indicator_loss}
    \end{equation}
    \normalsize
    where $\ell_{c,e}^{(t)}$ is defined in Eq.~\eqref{eq:loss} and $\alpha_L > 0$ is a hyperparameter that scales the impact of the loss.

\end{itemize}
The client-expert quality score $Q(c,e)$ is then updated using an Exponential Moving Average (EMA):
\small
\begin{equation} \label{eq:update_q}
    Q_{t}(c,e) = (1-\beta) \cdot Q_{t-1}(c,e) + \beta \cdot s_{c,e}^{(t)},
\end{equation}
\normalsize
where $\beta \in (0,1]$ is the learning rate for the EMA (e.g., $\beta=0.1$). For $e \notin \mathcal{A}_c^{(t)}$, the value is simply carried over:  $Q_t(c,e)=Q_{t-1}(c,e)$. The initial scores are set to a constant (e.g., $Q_0(c,e) = 0.2$) for all client-expert pairs.

\subsection{Expert Utilization Balance Evaluation Metrics} \label{sec:eval_metrics}
    
While standard metrics like prediction accuracy measure the overall performance of the trained MoE model, they do not capture how balanced expert utilization is across different experts. To this end, we propose specific metrics as follows.
    
\textbf{Per-Expert Load Coefficient of Variation (CV):} This metric measures the \textit{relative} imbalance in expert utilization over a recent period. Let $W_e^{(t, w)}$ be the total load (measured as the sum of usage counts) processed by expert $e$ during a sliding window of the last $w$ communication rounds (from round $t-w+1$ to $t$). This can be calculated from the client feedback: $W_e^{(t, w)} = \sum_{r=t-w+1}^{t} \sum_{c: e \in \mathcal{A}_c^{(r)}} u_{c,e}^{(r)}$. The CV at round $t$ for window $w$ is calculated by
    \small
    \begin{equation} \label{eq:cv_metric}
        CV^{(t, w)} = \frac{\operatorname{std}_e(W_e^{(t, w)})}{\operatorname{mean}_e(W_e^{(t, w)})}.
    \end{equation}
    \normalsize
        
    A lower CV indicates better expert utilization balance, with $CV=0$ representing perfect balance within the window. If the mean load is zero (e.g., during initial rounds), we define CV to be zero. We expect algorithms enforcing balanced expert utilization to achieve lower CV values compared to purely quality-driven approaches.

\subsection{Load-Balanced Expert Assignment Algorithm} \label{sec:assignment_algo}
Given the client-expert fitness scores $Q_{(t-1)}$ computed from feedback in the previous round, the server employs our proposed optimization-based algorithm to optimally assign expert subsets $\calA_c^{(t)}$ for each client $c$. Unlike simple greedy approaches that select top-$k_c$ experts per client (leading to severe load imbalance), our algorithm formulation jointly optimizes client-expert specialization and system-wide load balancing under resource constraints.

\subsubsection{Formulation for Joint Optimization}

This approach formulates expert assignment as an optimization problem that maximizes total client-expert quality scores while enforcing per-client capacity and per-expert load balance constraints.

\begin{enumerate}
    \item \textbf{Optimization Problem:} Let $X_{c,e} \in \{0, 1\}$ indicate whether expert $e$ is assigned to client $c$ in round $t$. The objective is to maximize the total accumulated quality:
    \small
    \begin{equation} \label{eq:ilp_objective}
        \max \sum_{c=1}^C \sum_{e=1}^E Q_{t-1}(c, e) \cdot X_{c,e},
    \end{equation}
    \normalsize
    subject to:
    \begin{itemize}
        \item \textbf{Client Capacity Constraint:}
        \small
        \begin{equation} \label{eq:ilp_client_cap}
            \sum_{e=1}^E X_{c,e} = k_c \quad \forall c \in \{1, ..., C\}.
        \end{equation}
        \normalsize
        \item \textbf{Expert Load Upper Bound:}
        \small
        \begin{equation} \label{eq:ilp_expert_upper}
            \sum_{c=1}^C |\mathcal{D}_c| \cdot X_{c,e} \le \Gamma_{\text{new}}(e) \quad \forall e \in \{1, ..., E\}.
        \end{equation}
        \normalsize
        \item \textbf{Expert Load Lower Bound:}
        \small
        \begin{equation} \label{eq:ilp_expert_lower}
            \sum_{c=1}^C |\mathcal{D}_c| \cdot X_{c,e} \ge L_{\text{new}}(e) \quad \forall e \in \{1, ..., E\}.
        \end{equation}
        \normalsize
    \end{itemize}

    \item \textbf{Dynamic Bounds Construction:} Prior to the assignment, the server computes the target per-expert load $\tau^{(t)} = \left(\sum_{c=1}^C |\mathcal{D}_c| \cdot k_c\right) / E$, reflecting the ideal assignment distribution for round $t$. For each expert $e$, the round-wise load deviation is defined as
    \small
    \begin{equation}
        d_e^{(t-1)} = W_e^{(t-1)} - \tau^{(t-1)},
    \end{equation}
    \normalsize
    where $W_e^{(t-1)} = \sum_{c: e \in \mathcal{A}_c^{(t-1)}} |\mathcal{D}_c|$ is the actual load assigned to expert $e$ in the previous round. The historical deficit $D_e^{(t)}$ is updated using exponential smoothing by
    \small
    \begin{equation}
        D_e^{(t)} = (1 - \gamma) D_e^{(t-1)} + \gamma d_e^{(t-1)}.
    \end{equation}
    \normalsize
    The target load for round $t$ is then adjusted by $\text{TargetLoad}_e^{(t)} = \tau^{(t)} - \alpha_{\text{adj}}\, D_e^{(t)}$, and the dynamic load bounds are
    \small
    \begin{align}
        L_{\text{new}}^{(t)}(e) &= \max(0,\text{TargetLoad}_e^{(t)} - \delta_{\text{dev}}),\\
        \Gamma_{\text{new}}^{(t)}(e) &= \text{TargetLoad}_e^{(t)} + \delta_{\text{dev}},
    \end{align}
    \normalsize
    with $\delta_{\text{dev}} = \delta_{\text{ratio}}\,\tau^{(t)}$, where $\gamma$, $\alpha_{\text{adj}}$, and $\delta_{\text{ratio}}$ are hyperparameters.
    
    \item \textbf{Solution Method:} The optimization problem is adapted from prior work on the Generalized Assignment Problem (GAP) \cite{ross1975branch} with additional dynamic load-balancing constraints. It is solved by the server utilizing standard solvers accessed through the PuLP library. The optimal binary assignment matrix $\{X_{c,e}\}$ determines each client's selected expert subset: 
    $\mathcal{A}_c^{(t)} = \{e \mid X_{c,e} = 1\}$.
\end{enumerate}

{In summary}, solving Eq.~\eqref{eq:ilp_objective} produces an expert assignment that optimizes fitness scores, adheres to client capacity constraints, and maintains bounded load imbalance across clients.

\section{Experiments}

In this section, we empirically evaluate the performance of our proposed FLEX-MoE framework.
We aim to demonstrate its effectiveness in training MoE models in resource-constrained federated settings, focusing on model accuracy and expert load balancing.

\subsection{Experimental Setup}

We conduct experiments on three benchmark datasets: CIFAR-10 \cite{krizhevsky2009learning}, EMNIST \cite{deng2012mnist}, and GTSRB \cite{stallkamp2011german}, simulating both IID and non-IID data distributions across clients. To simulate data heterogeneity, we employ two standard non-IID distributions: Dirichlet (controlling label concentration across clients) and Class partition (limiting each client to a fixed number of classes). 
Our model architecture incorporates an MoE layer with a ResNet model as each expert.
The MoE layer consists of $E$ experts, and each client $c$ can load at most $k_c$ experts.
We train for $T_{\text{max}} = 100$ communication rounds with $R = 3$ local training rounds per client.
For the main experiments, the client-expert quality score $Q$ is calculated by using Eqs. ~\eqref{eq:indicator_acc}- \eqref{eq:indicator_loss}.
Key evaluation metrics include average test accuracy across clients and expert utilization balance evaluated by the CV (Eq.~\eqref{eq:cv_metric}) with window $W = T_\text{max}$ covering the entire training period.

\begin{table}[t]
\caption{Accuracy (\%) and CV across three datasets under IID and moderately non-IID (Dirichlet $\alpha{=}0.8$) settings ($C=20$, $E=8$, $k_c\in[2,6]$).}
\centering
\small
\setlength{\tabcolsep}{4pt}
\begin{tabular}{ll|cc|cc|cc}
\toprule
\multirow{2}{*}{\textbf{Ind.}} & \multirow{2}{*}{\textbf{Algo.}}
& \multicolumn{2}{c|}{\textbf{CIFAR10}}
& \multicolumn{2}{c|}{\textbf{EMNIST}}
& \multicolumn{2}{c}{\textbf{GTSRB}} \\
\cmidrule{3-8}
& & \textbf{Acc.} & \textbf{CV}
  & \textbf{Acc.} & \textbf{CV}
  & \textbf{Acc.} & \textbf{CV} \\
\midrule
\multicolumn{8}{c}{\textbf{IID}} \\
\midrule
-- & Random   & 72.03 & 0.027 & 97.48 & 0.029 & 57.34 & 0.032 \\
\midrule
Acc  & Greedy    & 78.44 & 0.310 & 97.21 & 0.279 & 92.79 & 0.137 \\
     & FLEX-MoE  & \textbf{79.48} & \textbf{0.003} & \textbf{98.33} & \textbf{0.004} & \textbf{98.00} & \textbf{0.005} \\
\midrule
Loss & Greedy    & 78.71 & 0.298 & 98.02 & 0.167 & \textbf{90.88} & 0.096 \\
     & FLEX-MoE  & \textbf{83.00} & \textbf{0.003} & \textbf{98.08} & \textbf{0.004} & 76.48 & \textbf{0.005} \\
\midrule
\multicolumn{8}{c}{\textbf{Non-IID (Dirichlet $\alpha{=}0.8$)}} \\
\midrule
-- & Random   & 56.70 & 0.028 & 83.26 & 0.031 & 47.31 & 0.043 \\
\midrule
Acc  & Greedy    & 62.26 & 0.209 & \textbf{86.18} & 0.219 & \textbf{68.03} & 0.179 \\
     & FLEX-MoE  & \textbf{64.52} & \textbf{0.003} & 83.88 & \textbf{0.004} & 59.34 & \textbf{0.016} \\
\midrule
Loss & Greedy    & 61.14 & 0.295 & 85.06 & 0.209 & 67.94 & 0.015 \\
     & FLEX-MoE  & \textbf{68.81} & \textbf{0.003} & \textbf{86.91} & \textbf{0.004} & \textbf{76.50} & \textbf{0.008} \\
\bottomrule
\end{tabular}
\label{tab:full_comparison_with_bar_compact}
\end{table}

\subsection{Baselines}
We compare against two baselines. \textbf{Random Assignment} randomly assigns $k_c$ experts to each client, ensuring balanced utilization but ignoring client-expert fit. \textbf{Greedy Top-$k_c$ Selection} follows the greedy top-$k$ strategy of prior federated MoE work \cite{dun2023fedjets,mei2024fedmoe} using our fitness metrics: given the quality matrix $Q_{t-1}\in\mathbb{R}^{C\times E}$, each client $c$ independently selects its top-$k_c$ experts, $\mathcal{A}_c^{(t)} = \operatorname{TopK}_{k_c}(\boldsymbol{q}_c)$, prioritizing personalization while ignoring global load balancing.

\subsection{Comparison of General Results}

We evaluate all three methods under IID, moderately non-IID, and class-partitioned data distributions, with two expert scoring indicators (Acc- and Loss-driven) across three datasets (Table~\ref{tab:full_comparison_with_bar_compact}). Clients operate under limited capacity.

\textbf{FLEX-MoE achieves the best accuracy in most cases.} 
Our proposed FLEX-MoE method achieves the highest accuracy in the majority of configurations, though Greedy occasionally outperforms it in certain scenarios (e.g., GTSRB IID with Loss-driven indicator). FLEX-MoE demonstrates consistently superior performance through its joint optimization of expert-client matching and load balancing.

\textbf{FLEX-MoE provides consistently superior expert load balancing.} 
Across all configurations, FLEX-MoE achieves the lowest CV in expert usage, indicating nearly uniform expert training and superior expert utilization balance compared to all baselines. In contrast, Greedy often causes significant imbalance (e.g., CV exceeding 0.2 in some cases) as it prioritizes high-quality experts without considering global load distribution, leading to amplified expert usage skew.

\textbf{Computational overhead remains practical.} While the Greedy baseline adds virtually no runtime cost compared to Random, our proposed FLEX-MoE incurs a modest 6\% training time increase (24.3 vs. 25.8 minutes on dual NVIDIA A100 GPUs for CIFAR-10 IID) for significantly improved load balancing, making it practical for federated training.

\subsection{When Is Load Balancing Most Helpful?}
We now examine when load-balanced expert assignment is most beneficial, focusing on scenarios with strong data skew. Building on our earlier setup ($C=20$, $E=8$, $k_c \in [2,6]$), we evaluate two more challenging non-IID conditions: (1) 2-Class partition and (2) Dirichlet $\alpha = 0.1$.

\begin{table}[t]
\centering
\footnotesize
\caption{Performance (Acc.\ in \%) under severely skewed non-IID data on CIFAR-10 ($C=20$, $E=8$, $k_c\in[2,6]$).}
\label{tab:non_iid_extreme}
\setlength{\tabcolsep}{2.5pt}
\begin{tabular}{ll|cc|cc}
\toprule
\multirow{2}{*}{\textbf{Ind.}} & \multirow{2}{*}{\textbf{Algo.}}
& \multicolumn{2}{c|}{\textbf{2-Class partition}}
& \multicolumn{2}{c}{\textbf{Dirichlet $\alpha{=}0.1$}} \\
\cmidrule{3-6}
& & \textbf{Acc.} & \textbf{CV}
  & \textbf{Acc.} & \textbf{CV} \\
\midrule
-- & Random & 7.19 & 0.026 & 23.42 & 0.033 \\
\midrule
Acc  & Greedy   & 51.28 & 0.192 & 52.28 & 0.186 \\
     & FLEX-MoE & \textbf{61.00} & \textbf{0.003} & \textbf{67.60} & \textbf{0.002} \\
\midrule
Loss & Greedy   & 32.53 & 0.245 & 46.17 & 0.142 \\
     & FLEX-MoE & \textbf{59.44} & \textbf{0.003} & \textbf{55.75} & \textbf{0.003} \\
\bottomrule
\end{tabular}
\end{table}

\textbf{Load balancing is most helpful under severe data skew.}
Table~\ref{tab:non_iid_extreme} shows that FLEX-MoE consistently improves both accuracy and load balance over Greedy and Random in the 2-Class and Dirichlet $\alpha=0.1$ settings. Greedy can identify locally useful experts, but it repeatedly concentrates clients on a small subset of experts, yielding large CV values. FLEX-MoE keeps the assignment nearly balanced while preserving client-expert matching, which prevents unstable expert updates and improves accuracy when client data are highly personalized.

\textbf{FLEX-MoE's load balancing flexibility.} The parameter $\delta_{\text{ratio}}$ controls the deviation bounds around target expert loads, enabling a concise trade-off between expert specialization and load balance. Based on a validation set, we set $\delta_{\text{ratio}}=0.5$, which yields the best trade-off in our experiments.

\FloatBarrier
\section{Conclusion}
In this work, we proposed FLEX-MoE, a federated MoE framework that addresses the critical challenge of expert assignment in next-generation edge computing under resource constraints while ensuring system-wide load balancing. Our framework introduces a client-expert fitness score mechanism and a load-balanced expert assignment algorithm that jointly optimizes client-expert specialization and balanced expert utilization. Our experimental results demonstrate that load balancing is essential for effective federated MoE training, particularly under non-IID data distributions.
Our approach consistently achieves superior load balance while maintaining competitive accuracy, highlighting the importance of global optimization in federated MoE deployment for future AI-native wireless communications.

\FloatBarrier
\bibliographystyle{IEEEtran}
\bibliography{bib}

\end{document}